\documentclass[lettersize,journal,sort,compress]{IEEEtran}
\usepackage{amsmath,amssymb,amsfonts,bm}
\usepackage{algorithmic}
\usepackage{algorithm}
\usepackage{graphicx}
\usepackage{array}
\usepackage[caption=false,font=normalsize,labelfont=sf,textfont=sf]{subfig}
\usepackage{textcomp}
\usepackage{stfloats}
\usepackage{url}
\usepackage{verbatim}
\usepackage{xcolor}
\usepackage[colorlinks=true, allcolors=blue]{hyperref}
\usepackage{float}
\usepackage{booktabs}
\usepackage{multirow}
\usepackage{multirow}
\usepackage{pifont}
\usepackage{colortbl}
\usepackage{cite}
\title{\LARGE \bf Reparameterizable Dual-Resolution Network for Real-time Semantic Segmentation}

\author{
Guoyu Yang, Yuan Wang, Daming Shi* 

\thanks{This work is supported by Natural Science Foundation China (NSFC) No. 62372301 and No. U22A2097. (Corresponding author: Daming Shi.)}

\thanks{G. Yang and D. Shi are with the College of Computer Science and Software Engineering, Shenzhen University, Shenzhen 518060, China (e-mail: gyyang@zjut.edu.cn; dshi@szu.edu.cn).}

\thanks{Y. Wang is with the College of Computer Science and Technology, Zhejiang University of Technology, Hangzhou 310023, China (e-mail: sherlockwang@zjut.edu.cn).}
}

\begin{document}

\maketitle
\thispagestyle{empty}
\pagestyle{empty}

\begin{abstract}

Semantic segmentation plays a key role in applications such as autonomous driving and medical image. Although existing real-time semantic segmentation models achieve a commendable balance between accuracy and speed, their multi-path blocks still affect overall speed. To address this issue, this study proposes a Reparameterizable Dual-Resolution Network (RDRNet) dedicated to real-time semantic segmentation. Specifically, RDRNet employs a two-branch architecture, utilizing multi-path blocks during training and reparameterizing them into single-path blocks during inference, thereby enhancing both accuracy and inference speed simultaneously. Furthermore, we propose the Reparameterizable Pyramid Pooling Module (RPPM) to enhance the feature representation of the pyramid pooling module without increasing its inference time. Experimental results on the Cityscapes, CamVid, and Pascal VOC 2012 datasets demonstrate that RDRNet outperforms existing state-of-the-art models in terms of both performance and speed. The code is available at \url{https://github.com/gyyang23/RDRNet}.

\end{abstract}


\section{Introduction}

Semantic segmentation is a crucial task within the computer vision field, requiring the assignment of each pixel in an image to a specific semantic category. This technology is instrumental across various applications, including autonomous driving~\cite{cai2021multi}, medical image analysis~\cite{qi2023mdf}, and environmental monitoring~\cite{zhang2021lcu}. The performance of semantic segmentation models has greatly improved with the development of deep learning technology. However, there are still several challenges that need to be addressed, such as the model's inability to quickly analyze images. This limitation prohibits the direct deployment of the model to downstream tasks such as autonomous driving. Therefore, further research and improvements in semantic segmentation algorithms remain highly significant.

In the pursuit of meeting real-time or mobile requirements, numerous real-time semantic segmentation models~\cite{romera2017erfnet, yu2018bisenet, yu2021bisenet, pan2022deep, xu2023pidnet} have been proposed in the past. These models are characterized by fewer parameters, reduced computation, and rapid inference speed. One of the earliest models, ERFNet~\cite{romera2017erfnet}, achieved parameter and computation reduction by redesigning the ResNet~\cite{he2016deep} block using 1D convolution kernels and skip connections. However, the encoder-decoder architecture of ERFNet limits its ability to effectively learn spatial information from high-resolution features. In response to this limitation, BiSeNetV1~\cite{yu2018bisenet} introduced a two-branch architecture, with one branch focusing on spatial detail learning and the other on deep semantic information. While BiSeNetV1\&V2~\cite{yu2018bisenet, yu2021bisenet} have attained a favorable balance between speed and accuracy, the emergence of DDRNet~\cite{pan2022deep} and PIDNet~\cite{xu2023pidnet} has further improved the model's speed and accuracy, with the latter even matching or surpassing high-performance segmentation models in accuracy.

\begin{figure}[t]
\centering
\includegraphics[width=0.99\linewidth]{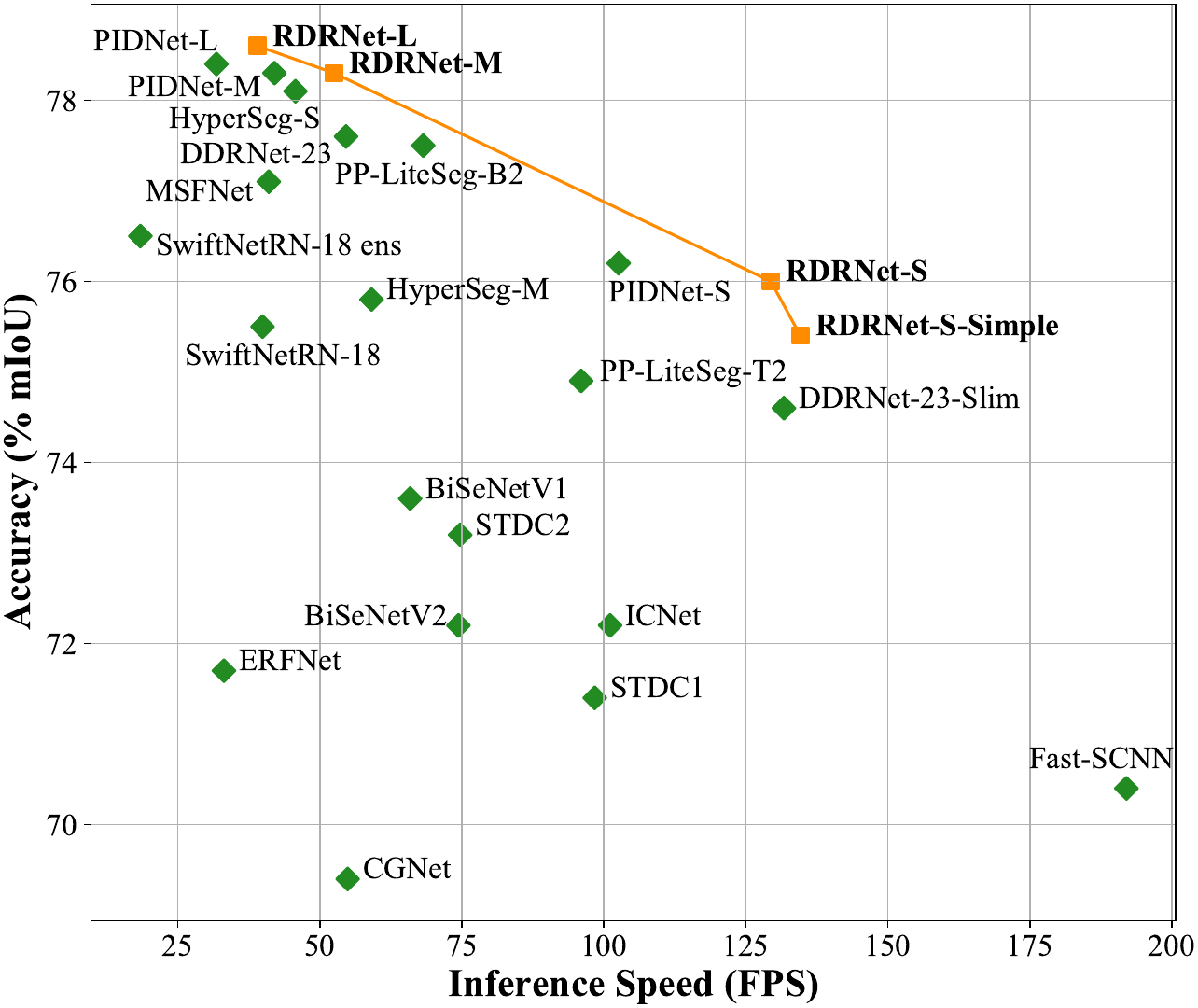}
\caption{The trade-off between inference speed and accuracy for real-time semantic segmentation models on the Cityscapes test set. Orange rectangles refer to our models, while green diamonds represent others.}
\label{0}
\end{figure}

Although these models~\cite{pan2022deep, xu2023pidnet} have achieved success in terms of speed and accuracy, the use of residual blocks~\cite{he2016deep} in the network architecture undoubtedly hinders the inference speed. While the residual block resolves the issue of gradient vanishing and exploding in deep neural networks, its residual connection increases computational costs and memory usage, ultimately impacting the inference speed. In comparison, single-path blocks~\cite{simonyan2014very}, owing to their unidirectional nature, are better suited for real-time segmentation. Multi-path blocks excel in training, but are not suitable for inference, whereas single-path blocks exhibit the opposite characteristics. Therefore, influenced by the aforementioned, we aim to train with multi-path blocks and infer with single-path blocks. While this way has been seldom explored in the realm of real-time semantic segmentation, in classification tasks, certain studies~\cite{ding2019acnet, ding2021diverse, ding2021repvgg} incorporate multi-path blocks during model training and subsequently reparameterize them into single-path blocks during inference.

In this study, drawing inspiration from previous research~\cite{ding2021repvgg, chen2024vanillanet}, we present a Reparameterizable Dual-Resolution Network (RDRNet) for real-time semantic segmentation tasks based on the two-branch architecture. Following the extraction of shallow features, RDRNet directs these features into two distinct branches. One branch focuses on learning deep semantic information (semantic branch), while the other delves into acquiring spatial detail information (detail branch). Notably, RDRNet doesn't solely fuse the output at the end of the two branches. Rather, it facilitates multiple interactions between the branches to achieve a more effective information fusion. We redesigned the ResNet block by eliminating redundant kernels, retaining only a singular $3 \times 3$ kernel, and introducing a new path with two $1 \times 1$ kernels. Importantly, the residual path remains unaltered. Throughout the training phase, the RDRNet block adheres to this configuration. However, during inference, it undergoes reparameterization, consolidating the three paths into a single-path with a $3 \times 3$ convolution kernel. Furthermore, recognizing the rapid but comparatively weaker performance of PAPPM~\cite{xu2023pidnet} compared to DAPPM~\cite{pan2022deep}, we propose a new pyramid pooling module: Reparameterizable Pyramid Pooling Module (RPPM). RPPM is able to learn more feature representations by adding a $3 \times 3$ grouped convolution in parallel to PAPPM. At inference time, this convolution is merged with another grouped convolution to generate a new $3 \times 3$ grouped convolution, ensuring that RPPM does not introduce additional computational burden. Notably, RPPM achieves performance similar to DAPPM while matching the speed of PAPPM.

To evaluate the performance and real-time capabilities of our model, we conducted experiments on three datasets: Cityscapes~\cite{cordts2016cityscapes}, CamVid~\cite{brostow2009semantic}, and Pascal VOC 2012~\cite{everingham2010pascal}. The experimental results demonstrate that RDRNet achieves the optimal balance between segmentation accuracy and inference speed, as shown in Figure~\ref{0}. Compared to other state-of-the-art real-time semantic segmentation models, our model exhibits superior performance and faster speed. In addition, we conducted ablation experiments to demonstrate the effectiveness of the proposed RPPM. 

The main contributions are summarized as follows:
\begin{itemize}
\item By leveraging the training advantages of multi-path blocks to enhance model's performance and reparameterizing multi-path blocks into single-path blocks during inference to boost inference speed, we propose a novel model termed Reparameterizable Dual-Resolution Network (RDRNet) for real-time semantic segmentation.
\item To enable the pyramid pooling module to learn richer feature representations without increasing inference time, we introduce the Reparameterizable Pyramid Pooling Module (RPPM).
\item Experiments on three public datasets demonstrate that the proposed model exhibits superior performance and faster inference speed compared to other state-of-the-art real-time semantic segmentation models.
\end{itemize}

\section{Related Work}

Semantic segmentation aims to assign each pixel in an image to its corresponding semantic class. Furthermore, semantic segmentation can be divided into high-performance semantic segmentation and real-time semantic segmentation.

\subsection{High-performance Semantic Segmentation}

High-performance semantic segmentation refers to the semantic segmentation of images on the premise of ensuring the quality and accuracy of segmentation. Being the earliest deep learning segmentation model, FCN~\cite{shelhamer2017fully} accomplishes end-to-end pixel-level semantic segmentation by substituting the fully connected layers in the traditional CNN model with convolutional layers. However, the excessive utilization of downsampling operations in FCN leads to the loss of spatial details in the feature map. To address this issue and enhance the receptive field without compromising spatial resolution, the DeepLab family~\cite{chen2017rethinking, chen2018encoder} integrates dilated convolutions~\cite{YuKoltun2016} with varying dilation rates into the network, deviating from the conventional convolutions. Additionally, in an effort to aggregate context information from multi-scale features and augment the network's capability to capture global information, PSPNet~\cite{zhao2017pyramid} introduces the Pyramid Pooling Module. In contrast to methodologies that capture context through multi-scale feature fusion, DANet~\cite{fu2019dual} employs a dual-attention mechanism within the network to adaptively merge local features with their global dependencies. In recent years, there has been significant development in transformers, leading to the emergence of several segmentation models~\cite{xie2021segformer, cheng2021per, cheng2022masked} based on the transformer structure. These models leverage the self-attention mechanism of transformers to effectively capture long-distance dependencies, resulting in notable improvements in performance for semantic segmentation tasks.

\subsection{Real-time Semantic Segmentation}

Real-time semantic segmentation refers to the semantic segmentation of the image under the premise of ensuring the segmentation speed. This approach will usually adopt a lightweight network architecture or improve the segmentation speed through techniques such as model compression and acceleration. Based on the architecture of real-time semantic segmentation models, the model can be further divided into two types: encoder-decoder architecture and multi-branch architecture.

\paragraph*{Encoder-decoder Architecture}
ERFNet~\cite{romera2017erfnet} introduces a novel block that utilizes a 1D convolution kernel and skip connection to reduce parameters and computation. Similarly, CGNet~\cite{wu2020cgnet} proposes the Context Guided block, which simultaneously learns local features and surrounding context features while further improving them with the global context. To improve inference speed and reduce parameters, CGNet uses channel-wise convolutions in the Context Guided block and carefully designs the entire network. STDC~\cite{fan2021rethinking} recognizes the time-consuming nature of adding an additional branch to compensate for the lack of spatial information in BiSeNet~\cite{yu2018bisenet}. To address this issue, STDC proposes a Detail Aggregation Module that preserves the spatial information of low-level features in a single-branch manner. In addition, PP-LiteSeg~\cite{peng2022pp} optimizes the decoder and introduces a Flexible and Lightweight Decoder to reduce computational overhead. 

\begin{figure*}[htbp]
\centering
\includegraphics[width=0.98\linewidth]{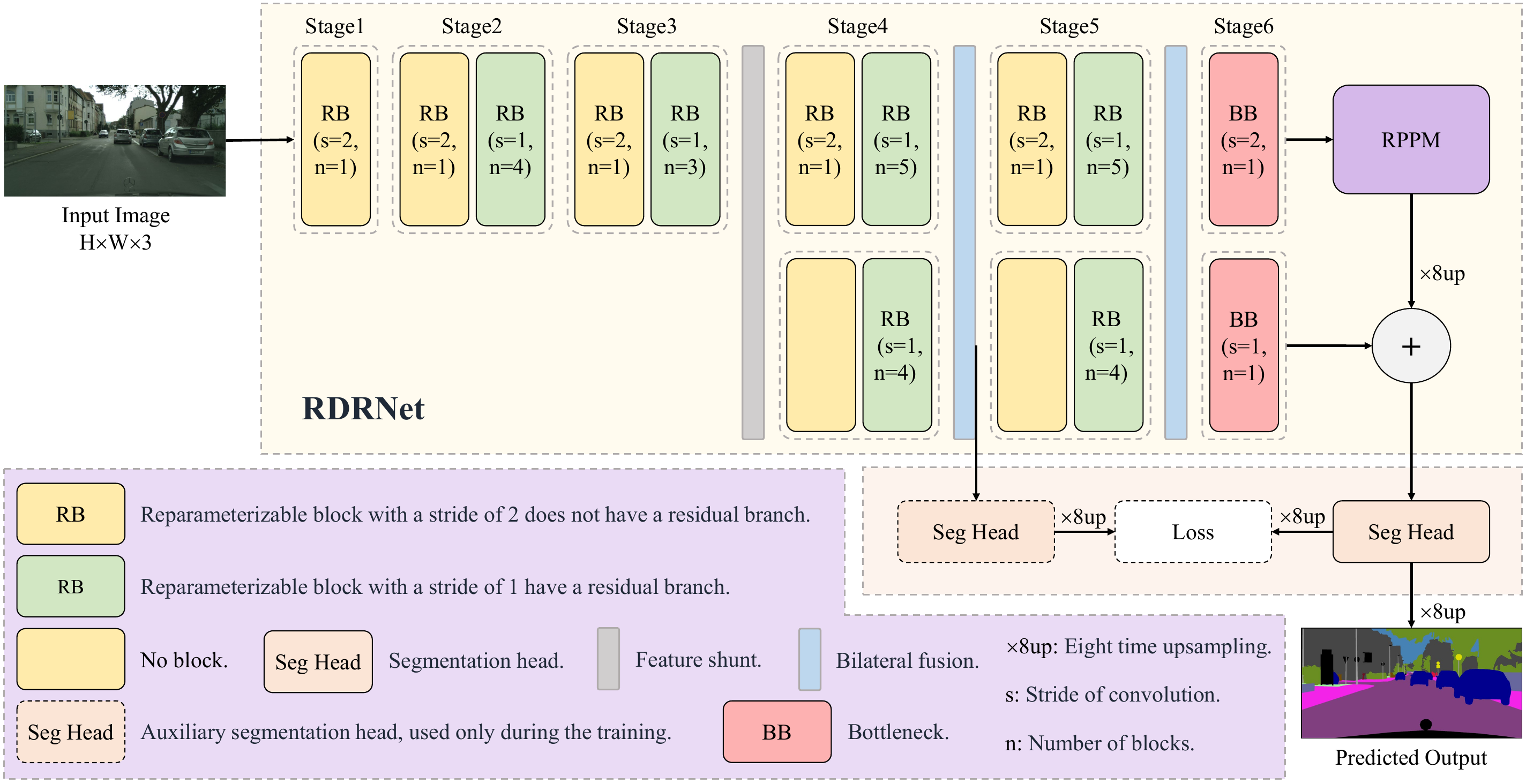}
\caption{Overall architecture of RDRNet. Following the feature shunting, the upper branch belongs to the semantic branch, while the lower branch belongs to the detail branch. RPPM refers to the proposed pyramid pooling module.}
\label{architecture}
\end{figure*}

\begin{figure}[t]
\centering
\includegraphics[width=0.95\linewidth]{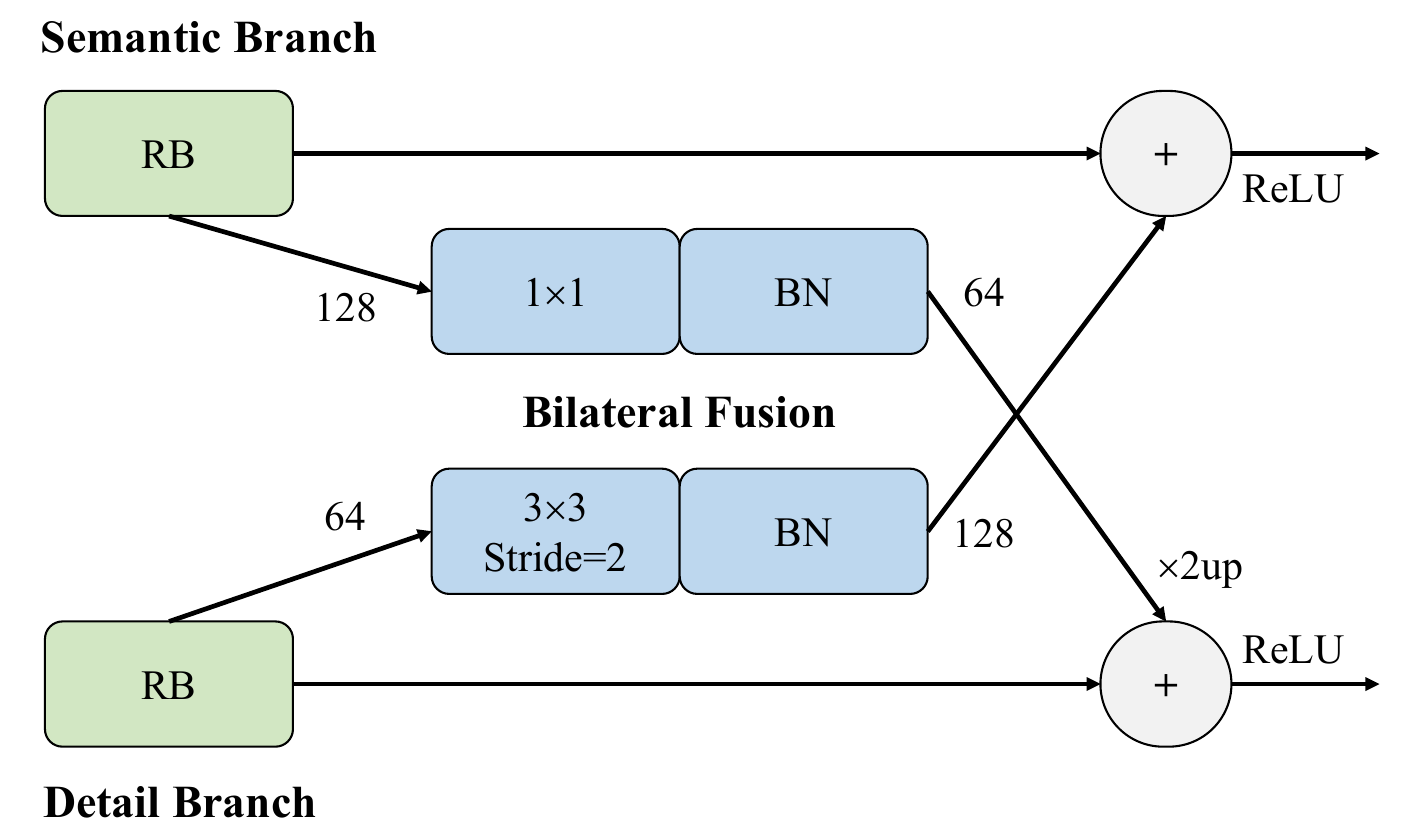}
\caption{The details of bilateral fusion in RDRNet. The bilateral fusion in the diagram corresponds to that after stage 4, with a similar fusion following stage 5. The key difference lies in the varying input/output channel numbers and upsampling rates.}
\label{bilateral fusion}
\end{figure}

\paragraph*{Multi-branch Architecture}
The encoder-decoder architecture model typically employs skip connections to connect the feature maps between the encoder and the decoder, thereby utilizing spatial details. In contrast, the multi-branch architecture model incorporates additional branches in the network, enabling the learning of spatial details. ICNet~\cite{zhao2018icnet} integrates multiresolution branches guided by appropriate labels. BiSeNetV1\&V2~\cite{yu2018bisenet, yu2021bisenet} introduced a two-branch architecture, with one branch dedicated to learning deep semantic information and the other branch focused on learning spatial information. It is important to note that these two branches do not share weights. In contrast, ~\cite{poudel2019fast, pan2022deep, xu2023pidnet} shares some backbone low-level weights. After extracting a certain shallow feature, Fast-SCNN~\cite{poudel2019fast} splits the feature into two branches: one to preserve the feature and the other to extract the global feature. DDRNet~\cite{pan2022deep} performs multiple bilateral fusion between the two branches to efficiently merge information, while introducing a Deep Aggregation Pyramid Pooling Module (DAPPM) that sequentially merges the extracted multi-scale context information. PIDNet~\cite{xu2023pidnet} proposes a three-branch architecture, utilizing three branches to parse detailed information, deep semantic information, and boundary information. Additionally, PIDNet introduces the Parallel Aggregation Pyramid Pooling Module (PAPPM), which is a modified version of DAPPM designed to enable parallel computing.

Although models based on the encoder-decoder architecture aim to strengthen their ability to learn spatial details, there is still a gap compared to multi-branch architecture models. This gap is reflected in the trade-off between accuracy and speed. Therefore, we implement our RDRNet based on a two-branch architecture. Existing real-time semantic segmentation models, whether based on encoder-decoder or multi-branch architecture, are hindered by the use of multi-path blocks, which limits the model's inference speed. In contrast to these models, our RDRNet employs multi-path blocks during training to ensure its learning capability. However, during inference, RDRNet reparameterizes multi-path blocks into a single-path block to enhance its speed without sacrificing accuracy.

\section{Method}

In this section, we initially present the overarching framework of the proposed Reparameterizable Dual-Resolution Network, followed by a detailed explanation of the Reparameterizable Block and the Reparameterizable Pyramid Pooling Module. Finally, we elaborate on the loss employed during the model's training phase.

\begin{table*}[htbp]
\renewcommand{\arraystretch}{1.3}
\caption{Detailed architectures of RDRNet-S and RDRNet-L. `[RB, 32, 2] $\times$ 1' indicates `[block used, output dimension, stride] $\times$ number'.}
\centering
\resizebox{\linewidth}{!}
{
\begin{tabular}{c|c|cc|cc} 
\specialrule{0.2em}{0em}{0.3em}
\textbf{Stage}           & \textbf{Output}                                   & \multicolumn{2}{c|}{\textbf{RDRNet-S}}                                                           & \multicolumn{2}{c}{\textbf{RDRNet-L}}                                                             \\
\specialrule{0.05em}{0em}{0em}
Input                    & $1024 \times 2048$                                &                 &                                                                                &                &                                                                                  \\
\specialrule{0.05em}{0em}{0em}
$S_{1}$                  & $512 \times 1024$                                 & \multicolumn{2}{c|}{[RB, 32, 2] $\times$ 1}                                                      & \multicolumn{2}{c}{[RB, 64, 2] $\times$ 1}                                                        \\
\specialrule{0.05em}{0em}{0em}
\multirow{2}*{$S_{2}$}   & \multirow{2}*{$256 \times 512$}                   & \multicolumn{2}{c|}{[RB, 32, 2] $\times$ 1}                                                      & \multicolumn{2}{c}{[RB, 64, 2] $\times$ 1}                                                        \\
                         &                                                   & \multicolumn{2}{c|}{[RB, 32, 1] $\times$ 4}                                                      & \multicolumn{2}{c}{[RB, 64, 1] $\times$ 6}                                                        \\
\specialrule{0.05em}{0em}{0em}
\multirow{2}*{$S_{3}$}   & \multirow{2}*{$128 \times 256$}                   & \multicolumn{2}{c|}{[RB, 64, 2] $\times$ 1}                                                      & \multicolumn{2}{c}{[RB, 128, 2] $\times$ 1}                                                       \\
                         &                                                   & \multicolumn{2}{c|}{[RB, 64, 1] $\times$ 3}                                                      & \multicolumn{2}{c}{[RB, 128, 1] $\times$ 5}                                                       \\
\specialrule{0.05em}{0em}{0em}
\multirow{3}*{$S_{4}$}   & \multirow{3}*{$64 \times 128, 128 \times 256$}    & \multicolumn{1}{c|}{[RB, 128, 2] $\times$ 1}   & \multirow{2}*{[RB, 64, 1] $\times$ 4}           & \multicolumn{1}{c|}{[RB, 256, 2] $\times$ 1}   & \multirow{2}*{[RB, 128, 1] $\times$ 6}           \\
                         &                                                   & \multicolumn{1}{c|}{[RB, 128, 1] $\times$ 5}   &                                                 & \multicolumn{1}{c|}{[RB, 256, 1] $\times$ 7}   &                                                  \\ \cline{3-6}
                         &                                                   & \multicolumn{2}{c|}{Bilateral Fusion}                                                            & \multicolumn{2}{c}{Bilateral Fusion}                                                              \\
\specialrule{0.05em}{0em}{0em}
\multirow{3}*{$S_{5}$}   & \multirow{3}*{$32 \times 64, 128 \times 256$}     & \multicolumn{1}{c|}{[RB, 256, 2] $\times$ 1}   & \multirow{2}*{[RB, 64, 1] $\times$ 4}           & \multicolumn{1}{c|}{[RB, 512, 2] $\times$ 1}   & \multirow{2}*{[RB, 128, 1] $\times$ 6}           \\
                         &                                                   & \multicolumn{1}{c|}{[RB, 256, 1] $\times$ 5}   &                                                 & \multicolumn{1}{c|}{[RB, 512, 1] $\times$ 7}   &                                                  \\ \cline{3-6}
                         &                                                   & \multicolumn{2}{c|}{Bilateral Fusion}                                                            & \multicolumn{2}{c}{Bilateral Fusion}                                                              \\
\specialrule{0.05em}{0em}{0em}
\multirow{1}*{$S_{6}$}   & \multirow{1}*{$16 \times 32, 128 \times 256$}     & \multicolumn{1}{c|}{[BB, 512, 2] $\times$ 1}   & \multicolumn{1}{c|}{[BB, 128, 1] $\times$ 1}    & \multicolumn{1}{c|}{[BB, 1024, 2] $\times$ 2}   & \multicolumn{1}{c}{[BB, 256, 1] $\times$ 2}     \\
\specialrule{0.2em}{0.3em}{0em}
\end{tabular}
}
\label{tab_architecture}
\end{table*}

\begin{figure}[t]
\centering
\includegraphics[width=0.95\linewidth]{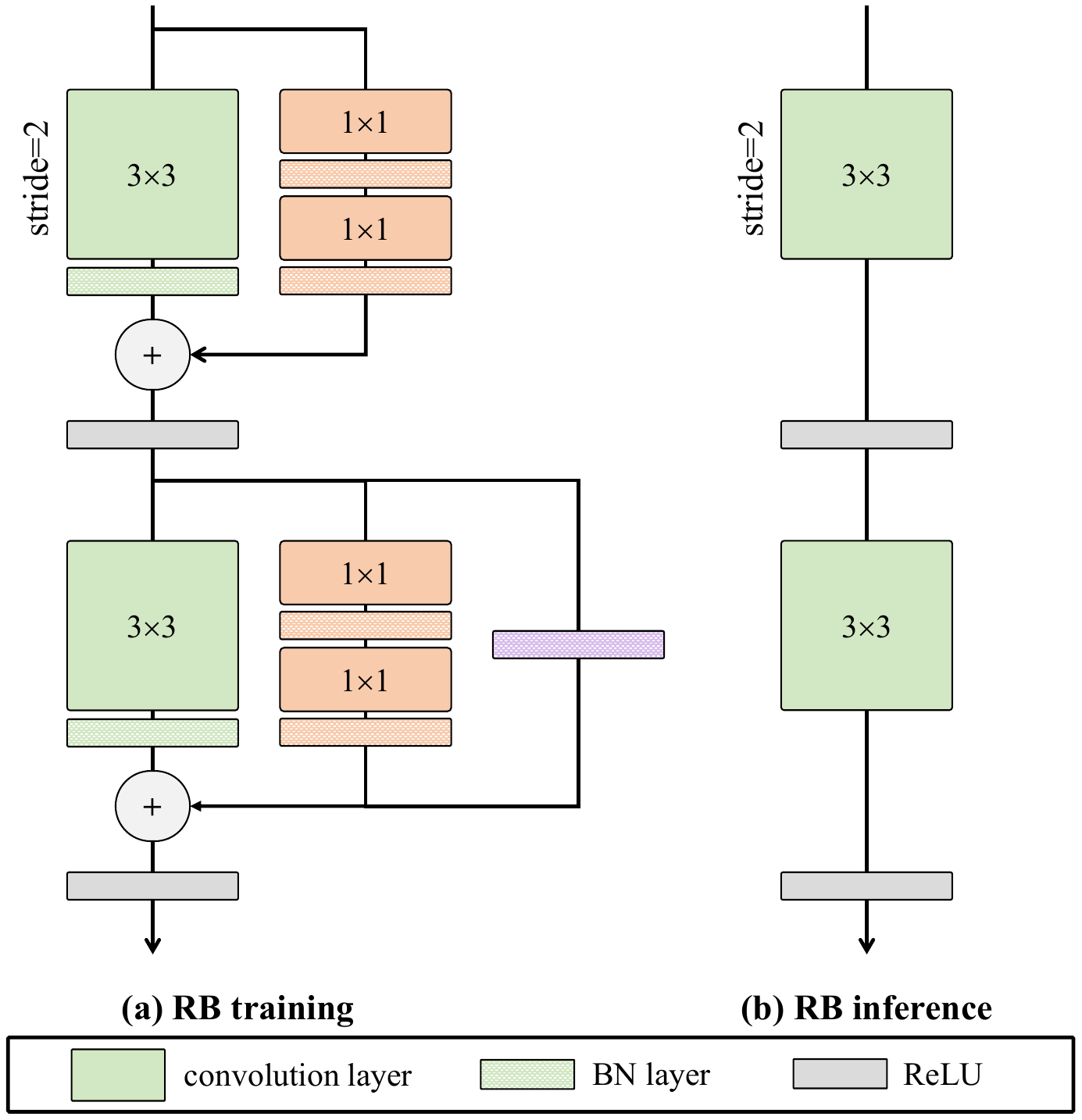}
\caption{Training and inference structure of the reparameterizable block. During non-downsampling, the training structure comprises three paths, while during downsampling, it reduces to two paths. In contrast to the training structure, the inference structure consists of only one path, where the convolutions are obtained through reparameterization from convolutions on other paths.}
\label{reparameterizable block}
\end{figure}

\subsection{Reparameterizable Dual-Resolution Network} \label{Reparameterizable Dual-Resolution Network}

As illustrated in Figure~\ref{architecture}, following the extraction of shallow features by RDRNet, the feature map is shunted into two branches. The upper branch, referred to as the semantic branch, is designed to learn deep semantic information. In contrast, the lower branch, termed the detail branch, is responsible for capturing spatial detail information. After feature extraction across multiple blocks in stage 4, the features from both branches are bilaterally fused, allowing the two branches to complement each other. A similar operation is performed in phase 5. After stage 6, the semantic branch features are fed into the pyramid pooling module to generate richer feature representations, while the detail branch features remain unchanged. Ultimately, the features from both branches are summed and passed to the segmentation head for prediction. During the training phase, an auxiliary segmentation head provides an additional loss function, facilitating the model's comprehensive learning of data features. Notably, since the auxiliary segmentation head is not utilized in the inference stage, it does not impact the model's inference speed and efficiency.

In previous studies, works such as DDRNet~\cite{pan2022deep} and PIDNet~\cite{xu2023pidnet} adopt ResNet~\cite{he2016deep} residual block as the basic block in their models. To eliminate the slower inference speed due to residual block, we use the proposed reparameterizable block as the basic block of RDRNet. In Figure~\ref{architecture}, it can be seen that the reparameterizable blocks occupy a major part of the whole model (stages 1 to 5), which is the main factor for the inference speed improvement of our RDRNet. When performing bilateral fusion, RDRNet will use $1 \times 1$ convolution to compress the features of the semantic branch, use bilinear interpolation for upsampling, and then added to the features of the detail branch. The features of the detail branch are expanded and downsampled using a $3 \times 3$ convolution (or two) with a stride of 2, and then added to the features of the semantic branch. Figure~\ref{bilateral fusion} illustrates the bilateral fusion process after stage 4. Assuming that $R_{S}$ and $R_{D}$ correspond to a series of reparameterizable blocks of the semantic branch and the detail branch, $T_{S \rightarrow D}$ and $T_{D \rightarrow S}$ denote the semantics-to-detail feature alignment operation and the detail-to-semantics feature alignment operation, respectively. After bilateral fusion, the $i$-th features of the semantic branch, denoted as $X^{i}_{S}$, and the $i$-th features of the detail branch, denoted as $X^{i}_{D}$, can be articulated as follows:
\begin{equation}
\label{eq1}
\begin{cases}
X^{i}_{S} = ReLU(R_{S}(X^{i-1}_{S}) + T_{D \rightarrow S}(R_{D}(X^{i-1}_{D})))\\
X^{i}_{D} = ReLU(R_{D}(X^{i-1}_{D}) + T_{S \rightarrow D}(R_{S}(X^{i-1}_{S})))
\end{cases}
\end{equation}

The segmentation head of our RDRNet is similar to DDRNet, comprising a $3 \times 3$ convolution followed by a $1 \times 1$ convolution. The $3 \times 3$ convolution is utilized to learn the features after combining the semantic features and detailed features, while also adjusting the channel dimension ($O_{c}$). The $1 \times 1$ convolution is employed to align the feature channels with the number of classes; for instance, if the number of classes is 19, the $1 \times 1$ convolution will adjust the feature from channel $O_{c}$ to 19.

\begin{figure}[t]
\centering
\includegraphics[width=0.9\linewidth]{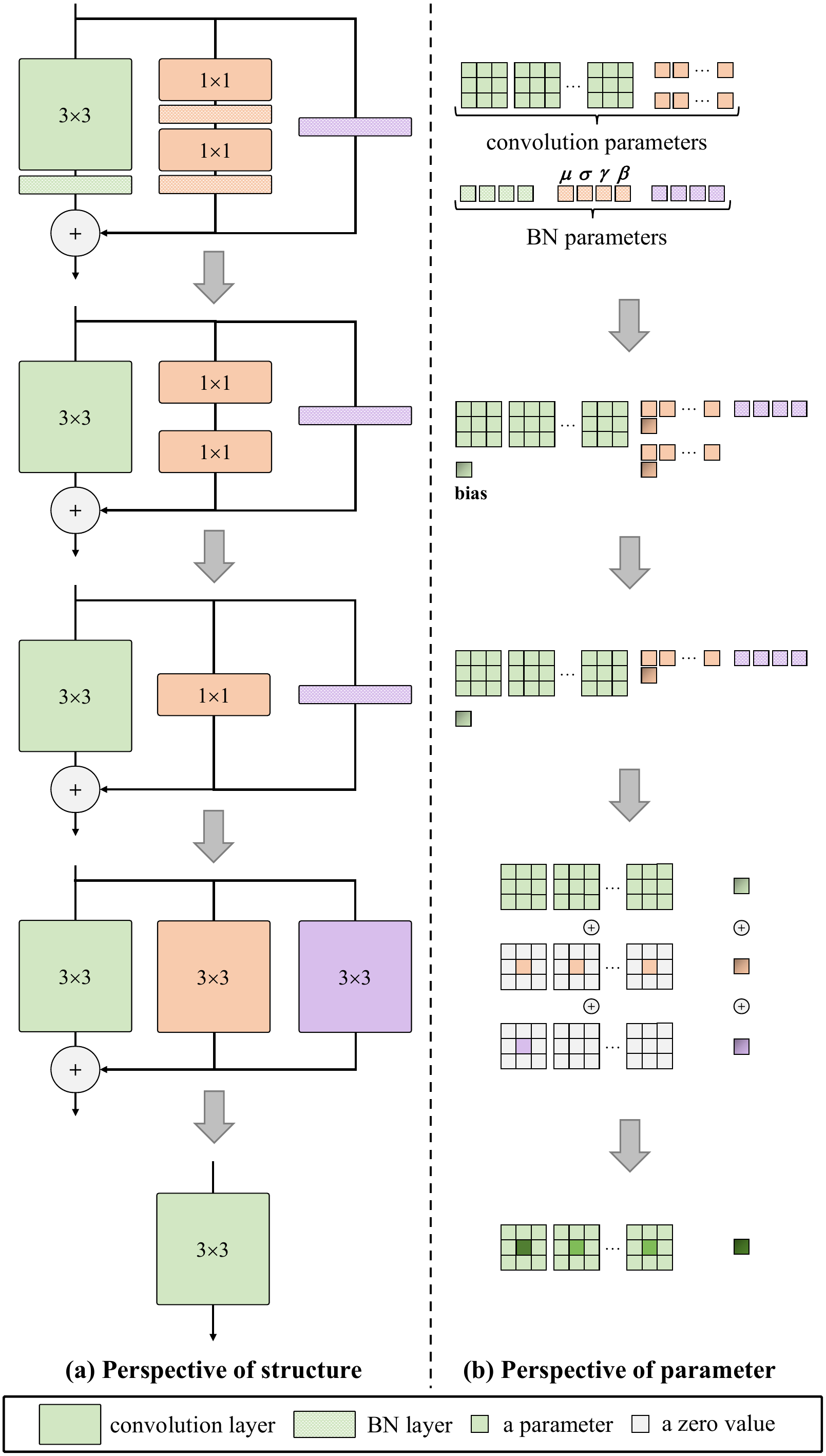}
\caption{The reparameterization process of RB from two different perspectives, namely, transitioning from a multi-path structure to a single-path structure.}
\label{reparameterization}
\end{figure}

We scale the depth and width of RDRNet, and construct a total of four different versions of RDRNet, namely RDRNet-S-Simple, RDRNet-S, RDRNet-M, and RDRNet-L. In Table~\ref{tab_architecture}, we present the detailed architectures of RDRNet-S and RDRNet-L. RDRNet-S-Simple and RDRNet-S share the same depth and width, but differ in the $O_{c}$ in the segmentation head, while RDRNet-M has double the width of RDRNet-S. We set the $O_{c}$ of each of the four models to 64, 128, 128, and 256, respectively.

\subsection{Reparameterizable Block}

As shown in Figure~\ref{reparameterizable block}, the Reparameterizable Block (RB) has three paths during training: the first path applies a $3 \times 3$ convolution, the second path applies two $1 \times 1$ convolutions, and the third path is the residual connection. The residual connection are removed when performing the downsampling operation. This is because the residual connection is used to maintain the input feature map, and the downsampling operation will halve the spatial resolution of the feature map, making the residual path no longer applicable. At inference time, RDRNet reparameterizes RB to a $3 \times 3$ convolution without loss of accuracy. Compared with the ResNet block, a path consisting of two $1 \times 1$ convolutions is added to RB. This path enables RB to learn more feature representations, which helps to improve the performance of the model.

When performing reparameterization, RB first merge the convolutional weight with the parameters of batch normalization (BN). Assuming that the scale factor, offset factor, mean, and variance in BN are $\gamma$, $\beta$, $\mu$, and $\sigma \in \mathbb{R}^{C_{out}}$, respectively, for a $k \times k$ convolution with input channel $C_{in}$ and output channel $C_{out}$, the weight and bias obtained by merging the convolution weight $W \in \mathbb{R}^{C_{out} \times (C_{in} \times k \times k)}$ with bias $B \in \mathbb{R}^{C_{out}}$ and BN are as follows:
\begin{equation}
\label{eq2}
W^{\prime} = \frac{\gamma}{\sigma} W, B^{\prime} = \frac{(B - \mu)\gamma}{\sigma} + \beta.
\end{equation}

As depicted in Figure~\ref{reparameterization}, after merging the convolution and BN, RB will combine the two $1 \times 1$ convolutions in series. Assuming two $1 \times 1$ convolutions represented by $W_{1 \times 1}^{(1)} \in \mathbb{R}^{C_{out1} \times (C_{in} \times 1 \times 1)}$ and $W_{1 \times 1}^{(2)} \in \mathbb{R}^{C_{out2} \times (C_{out1} \times 1 \times 1)}$, with input feature $x \in \mathbb{R}^{C_{in} \times H \times W}$ and output feature $y \in \mathbb{R}^{C_{out2} \times H^{\prime} \times W^{\prime}}$, the convolutions can be expressed as:
\begin{equation}
\label{eq3}
\begin{split}
y &= W_{1 \times 1}^{(2)} * (W_{1 \times 1}^{(1)} * x)  \\
  &= W_{1 \times 1}^{(2)} \cdot im2col_{2}(reshape(W_{1 \times 1}^{(1)} \cdot im2col_{1}(x))),
\end{split}
\end{equation}
where $*$ denotes the convolution operation, and $\cdot$ denotes matrix multiplication. The $im2col$ operator will transform the input $x$ into a two-dimensional matrix that corresponds to the shape of the convolution kernel. For example, $im2col_{1}$ will transform $x$ of shape $C_{in} \times H \times W$ into $X$ of shape $(C_{in} \times 1 \times 1) \times (H^{\prime} \times W^{\prime})$. The $reshape$ operator will transform the resulting matrix into a tensor (feature map). Since $W_{1 \times 1}^{(2)}$ is a $1 \times 1$ convolution kernel of shape $C_{out2} \times (C_{out1} \times 1 \times 1)$, and $W_{1 \times 1}^{(1)} \cdot im2col_{1}(x)$ has a shape of $C_{out1} \times (H^{\prime} \times W^{\prime})$, we conclude that:
\begin{equation}
\label{eq4}
\begin{split}
&~~~~ im2col_{2}(reshape(W_{1 \times 1}^{(1)} \cdot im2col_{1}(x)))~~~~~~~    \\
&= W_{1 \times 1}^{(1)} \cdot im2col_{1}(x).
\end{split}
\end{equation}
According to Equation~\ref{eq3} and Equation~\ref{eq4}, it can be further deduced as follows.
\begin{equation}
\label{eq5}
\begin{split}
y &= W_{1 \times 1}^{(2)} \cdot im2col_{2}(reshape(W_{1 \times 1}^{(1)} \cdot im2col_{1}(x))) \\
  &= W_{1 \times 1}^{(2)} \cdot W_{1 \times 1}^{(1)} \cdot im2col_{1}(x)                      \\
  &= (W_{1 \times 1}^{(2)} \cdot W_{1 \times 1}^{(1)}) \cdot im2col_{1}(x)                    \\
  &= W_{1 \times 1} * x,
\end{split}
\end{equation}
where $W_{1 \times 1}$ is the new convolution produced after merging. It follows that two $1 \times 1$ convolutions can be combined into a single $1 \times 1$ convolution without incurring any performance degradation. It is worth noting that Equation~\ref{eq4} holds only when the stride of the second $1 \times 1$ convolution is 1. When the RB performs downsampling, the stride of the first $1 \times 1$ convolution is set to 2, whereas that of the second $1 \times 1$ convolution is set to 1.

After merging the two concatenated $1 \times 1$ convolutions, the RB is left with a $3 \times 3$ convolution, a $1 \times 1$ convolution, and a residual connection. Subsequently, RB will reparameterize the $1 \times 1$ convolutions and residual connection into $3 \times 3$ convolutions, respectively. As illustrated in Figure~\ref{reparameterization}, it is intuitively apparent that the $1 \times 1$ convolution is a special case of the $3 \times 3$ convolution, where the central element has a non-zero weight value and the other elements have a weight value of 0. Therefore, RB can readily reparameterize the $1 \times 1$ convolution into a $3 \times 3$ convolution. Regarding the residual connection, RB will first construct a $1 \times 1$ convolution $W_{rc} \in \mathbb{R}^{C_{i} \times C_{j} \times 1 \times 1}$ $ (i \leq j \leq out)$ to supplant it, where the weight value is 1 if $i$ equals $j$ and 0 if $i$ is not equal to $j$. The $1 \times 1$ convolution is then reparameterized into a $3 \times 3$ convolution. Since now all three paths are $3 \times 3$ convolutions, RB can directly add the weights and biases of the three $3 \times 3$ convolutions to obtain a new $3 \times 3$ convolution.

Overall, RB not only benefits from the training advantages of multi-path blocks but also inherits the inference advantages of single-path blocks. This design not only enables the model to fully exploit the rich information and complex feature representations of multi-path blocks during training but also facilitates rapid image segmentation during the inference stage.

\begin{figure}[t]
\centering
\includegraphics[width=0.99\linewidth]{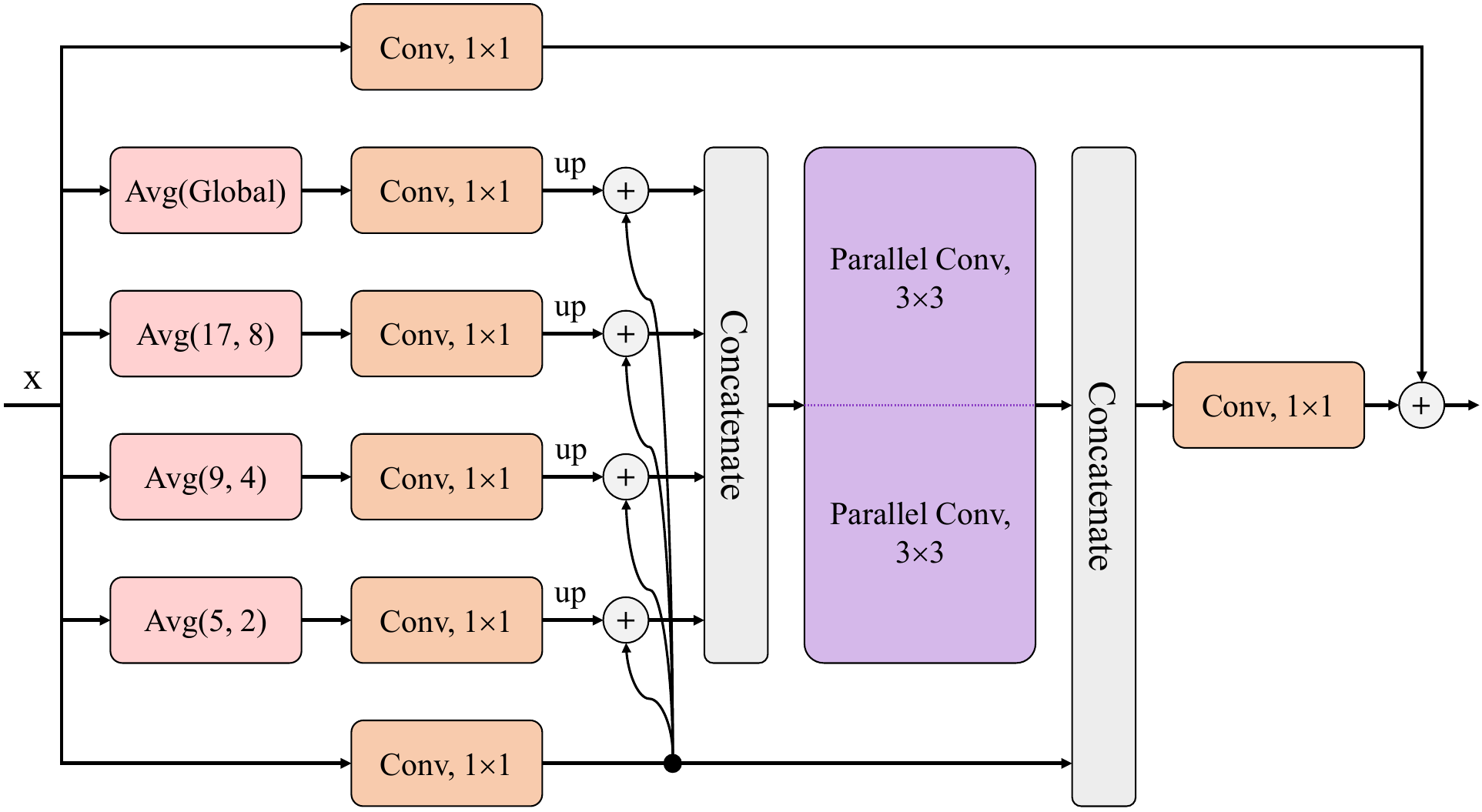}
\caption{The detailed structure of RPPM. Parallel computation is achieved through the utilization of grouped convolution.}
\label{RPPM}
\end{figure}

\subsection{Reparameterizable Pyramid Pooling Module}

The Pyramid Pooling Module (PPM)~\cite{zhao2017pyramid} is utilized to capture features of the image at multiple scales, aiding the model in comprehending and analyzing local and global information within the image. In the real-time semantic segmentation model, the proposed Deep Aggregation PPM (DAPPM)~\cite{pan2022deep} and Parallel Aggregation PPM (PAPPM)~\cite{xu2023pidnet} substantially enhance the model's performance. While PAPPM employs parallel processing based on DAPPM to augment module speed, the module's precision diminishes due to the omission of certain operations in parallelization. To achieve both the speed of PAPPM and the performance of DAPPM, we introduce Reparameterizable PPM (RPPM).

Figure~\ref{RPPM} illustrates the overall structure of RPPM, providing a clear visualization of the flow process from input features to output features, where parallel convolutions are essentially grouped convolutions. Given that the grouped convolution in PyTorch utilizes parallel computing technology, in scenarios where different convolutions operate on distinct inputs, we can concatenate disparate inputs and subsequently pass them to the grouped convolution, thereby harnessing parallel computing to enhance inference speed. In RPPM, we employ two parallel-structured $3 \times 3$ grouped convolutions to enable the module to learn richer feature representations, thereby improving performance. Notably, these two convolutions are only present during training, and at test time, RPPM reparameterizes them into a single, optimized grouped convolution. The process is similar to the RB reparameterization process. After training, RPPM firstly fuses two $3 \times 3$ grouped convolutions with their corresponding BN layers, and then merges the weights and biases of the two convolutions by element addition. Since the inference structure of RPPM is consistent with that of PAPPM, RPPM successfully maintains the inference efficiency of PAPPM while achieving a performance boost.

\subsection{Deep Supervision}

Previous studies~\cite{yu2021bisenet, pan2022deep, xu2023pidnet} have demonstrated that incorporating the auxiliary segmentation head during the model training phase can enhance segmentation performance without incurring additional inference costs. In RDRNet, we employ an additional auxiliary segmentation head during training, which is subsequently removed during testing. As illustrated in Figure~\ref{architecture}, this auxiliary segmentation head is situated after the stage 4 bilateral fusion of RDRNet, receiving feature from the detail branch. Following forward propagation of the model, the total loss can be expressed as:
\begin{equation}
\label{eq6}
L = L_{n} + {\alpha}L_{a}, 
\end{equation}
where $L_{n}$ and $L_{a}$ represent the normal loss and auxiliary loss, respectively, and $\alpha$ denotes the weight coefficient of the auxiliary loss, set to 0.4 in RDRNet. To effectively tackle imbalanced data and hard samples, we also adopt OHEM Cross Entropy as the loss function, consistent with previous work~\cite{xu2023pidnet}.

\begin{table*}[htbp]
\renewcommand{\arraystretch}{1.2}
\caption{Accuracy and speed analysis on the Cityscapes validation and test sets. `*' denotes results re-implemented based on MMSegmentation and speed-tested on our platform. It's worth noting that, for fair comparison, models with `*' and our model had convolution and BN fused during speed testing. ImageNet indicates whether the model was pre-trained on the ImageNet dataset.}
\centering
\resizebox{\linewidth}{!}
{
\begin{tabular}{lcccccccc} 
\specialrule{0.2em}{0em}{0.3em}
\multirow{2}*{\textbf{Method}}              & \multirow{2}*{\textbf{Resolution}}   & \multirow{2}*{\textbf{GPU}}  & \multirow{2}*{\textbf{FPS}}  & \multirow{2}*{\textbf{Params}}   & \multirow{2}*{\textbf{GFLOPs}}   & \multirow{2}*{\textbf{ImageNet}}   & \multicolumn{2}{c}{\textbf{mIoU (\%)}}    \\ \cmidrule{8-9}
                                &                                      &                              &                              &                                  &                                  &                                    & \textbf{val}          & \textbf{test}     \\
\specialrule{0.15em}{0.3em}{0.3em}
ERFNet*~\cite{romera2017erfnet}             & $1024 \times 2048$                   & RTX 3090                     & 33.1                         & 2.1 M                            & 117.0                            & \ding{55}                          & 72.5                  & 71.7              \\
\specialrule{0.05em}{0.3em}{0.3em}
ICNet*~\cite{zhao2018icnet}                 & $1024 \times 2048$                   & RTX 3090                     & 101.1                        & 24.9 M                           & 74.3                             & \ding{55}                          & 71.6                  & 72.2              \\
\specialrule{0.05em}{0.3em}{0.3em}
Fast-SCNN*~\cite{poudel2019fast}            & $1024 \times 2048$                   & RTX 3090                     & 192.0                        & 1.4 M                            & 7.3                              & \ding{55}                          & 71.0                  & 70.4              \\
\specialrule{0.05em}{0.3em}{0.3em}
MSFNet~\cite{si2019real}                    & $1024 \times 2048$                   & RTX 2080Ti                   & 41.0                         & -                                & 96.8                             & \ding{51}                          & -                     & 77.1              \\
\specialrule{0.05em}{0.3em}{0.3em}
SwiftNetRN-18~\cite{orsic2019defense}       & $1024 \times 2048$                   & GTX 1080Ti                   & 39.9                         & 11.8 M                           & 104.0                            & \ding{51}                          & 75.4                  & 75.5              \\
SwiftNetRN-18 ens~\cite{orsic2019defense}   & $1024 \times 2048$                   & GTX 1080Ti                   & 18.4                         & 24.7 M                           & 218.0                            & \ding{51}                          & -                     & 76.5              \\
\specialrule{0.05em}{0.3em}{0.3em}
CGNet*~\cite{wu2020cgnet}                   & $1024 \times 2048$                   & RTX 3090                     & 54.9                         & 0.5 M                            & 27.5                             & \ding{55}                          & 68.3                  & 69.4              \\
\specialrule{0.05em}{0.3em}{0.3em}
BiSeNetV1*~\cite{yu2018bisenet}             & $1024 \times 2048$                   & RTX 3090                     & 65.9                         & 13.3 M                           & 118.0                            & \ding{51}                          & 74.4                  & 73.6              \\
BiSeNetV2*~\cite{yu2021bisenet}             & $1024 \times 2048$                   & RTX 3090                     & 74.4                         & 3.4 M                            & 98.4                             & \ding{55}                          & 73.6                  & 72.2              \\
\specialrule{0.05em}{0.3em}{0.3em}
STDC1*~\cite{fan2021rethinking}             & $1024 \times 2048$                   & RTX 3090                     & 98.4                         & 8.3 M                            & 67.5                             & \ding{55}                          & 71.8                  & 71.4              \\
STDC2*~\cite{fan2021rethinking}             & $1024 \times 2048$                   & RTX 3090                     & 74.7                         & 12.3 M                           & 93.8                             & \ding{55}                          & 74.9                  & 73.2              \\
\specialrule{0.05em}{0.3em}{0.3em}
HyperSeg-M~\cite{nirkin2021hyperseg}        & $512 \times 1024$                    & RTX 3090                     & 59.1                         & 10.1 M                           & 7.5                              & \ding{51}                          & 76.2                  & 75.8              \\
HyperSeg-S~\cite{nirkin2021hyperseg}        & $768 \times 1536$                    & RTX 3090                     & 45.7                         & 10.2 M                           & 17.0                             & \ding{51}                          & 78.2                  & 78.1              \\
\specialrule{0.05em}{0.3em}{0.3em}
PP-LiteSeg-T2~\cite{peng2022pp}             & $768 \times 1536$                    & RTX 3090                     & 96.0                         & -                                & -                                & \ding{51}                          & 76.0                  & 74.9              \\
PP-LiteSeg-B2~\cite{peng2022pp}             & $768 \times 1536$                    & RTX 3090                     & 68.2                         & -                                & -                                & \ding{51}                          & 78.2                  & 77.5              \\
\specialrule{0.05em}{0.3em}{0.3em}
DDRNet-23-Slim*~\cite{pan2022deep}          & $1024 \times 2048$                   & RTX 3090                     & 131.7                        & 5.7 M                            & 36.3                             & \ding{55}                          & 76.3                  & 74.6              \\
DDRNet-23*~\cite{pan2022deep}               & $1024 \times 2048$                   & RTX 3090                     & 54.6                         & 20.3 M                           & 143.0                            & \ding{55}                          & 78.0                  & 77.6              \\
\specialrule{0.05em}{0.3em}{0.3em}
PIDNet-S*~\cite{xu2023pidnet}               & $1024 \times 2048$                   & RTX 3090                     & 102.6                        & 7.7 M                            & 47.3                             & \ding{55}                          & 76.4                  & 76.2              \\
PIDNet-M*~\cite{xu2023pidnet}               & $1024 \times 2048$                   & RTX 3090                     & 42.0                         & 28.7 M                           & 177.0                            & \ding{55}                          & 78.2                  & 78.3              \\
PIDNet-L*~\cite{xu2023pidnet}               & $1024 \times 2048$                   & RTX 3090                     & 31.8                         & 37.3 M                           & 275.0                            & \ding{55}                          & 78.8                  & 78.4              \\
\specialrule{0.05em}{0.3em}{0.3em}
\rowcolor{green!25}
RDRNet-S-Simple                             & $1024 \times 2048$                   & RTX 3090                     & 134.6                        & 7.2 M                            & 41.0                             & \ding{55}                          & 76.8                  & 75.4              \\
\rowcolor{green!25}
RDRNet-S                                    & $1024 \times 2048$                   & RTX 3090                     & 129.4                        & 7.3 M                            & 43.4                             & \ding{55}                          & 76.8                  & 76.0              \\
\rowcolor{green!25}
RDRNet-M                                    & $1024 \times 2048$                   & RTX 3090                     & 52.5                         & 26.0 M                           & 162.0                            & \ding{55}                          & 78.9                  & 78.3              \\
\rowcolor{green!25}
RDRNet-L                                    & $1024 \times 2048$                   & RTX 3090                     & 39.0                         & 36.9 M                           & 238.0                            & \ding{55}                          & 79.3                  & 78.6              \\
\specialrule{0.2em}{0.3em}{0em}
\end{tabular}
}
\label{tab_cityscapes}
\end{table*}

\section{Experiments} 

\subsection{Datasets}

\noindent\textbf{Cityscapes.} Cityscapes~\cite{cordts2016cityscapes} is widely used in the fields of urban scene understanding and autonomous driving. It encompasses 19 classes and comprises 5000 images, with 2975 allocated for training, 500 for validation, and 1525 for testing. These images undergo meticulous annotation and possess a resolution of $1024 \times 2048$ pixels.

\noindent\textbf{CamVid.} CamVid~\cite{brostow2009semantic} is the first video set with semantic labels. It encompasses 701 images, with 367 allocated for training, 101 for validation, and 233 for testing. Each image possesses a resolution of $720 \times 960$ pixels. The dataset features 32 class labels, although typically only 11 of them are employed for training and evaluation.

\noindent\textbf{Pascal VOC 2012.} Pascal VOC 2012~\cite{everingham2010pascal} is mainly used for tasks such as image classification, object detection, and image segmentation. It covers 20 classes and 1 background class. There are a total of 2913 images used for the semantic segmentation task, including 1464 images in the training set and 1449 images in the validation set. Unlike Cityscapes and CamVid, the resolution of these images is not fixed.

\begin{table}[t]
\renewcommand{\arraystretch}{1.1}
\setlength{\tabcolsep}{10.5pt}
\caption{Accuracy and speed analysis on the CamVid test set. All models in the table are results of our reimplementation based on MMSegmentation and were speed-tested on our platform (RTX 3090). During training, all models used Cityscapes weight, and during speed evaluation, convolution and BN were fused. ImageNet indicates whether the model was pre-trained on the ImageNet dataset.}
\centering
\resizebox{\linewidth}{!}
{
\begin{tabular}{lcccc} 
\specialrule{0.15em}{0em}{0.3em}
\textbf{Method}                     & \textbf{ImageNet}  & \textbf{FPS}           & \textbf{mIoU}    \\ 
\specialrule{0.1em}{0.3em}{0.3em}
ERFNet~\cite{romera2017erfnet}      & \ding{55}          & 71.2                   & 72.4             \\
\specialrule{0.03em}{0.3em}{0.3em}
ICNet~\cite{zhao2018icnet}          & \ding{55}          & 173.7                  & 65.7             \\
\specialrule{0.03em}{0.3em}{0.3em}
Fast-SCNN~\cite{poudel2019fast}     & \ding{55}          & 260.6                  & 66.2             \\
\specialrule{0.03em}{0.3em}{0.3em}
CGNet~\cite{wu2020cgnet}            & \ding{55}          & 99.5                   & 69.4             \\
\specialrule{0.03em}{0.3em}{0.3em}
BiSeNetV1~\cite{yu2018bisenet}      & \ding{51}          & 140.2                  & 73.3             \\
BiSeNetV2~\cite{yu2021bisenet}      & \ding{55}          & 156.9                  & 75.8             \\
\specialrule{0.03em}{0.3em}{0.3em}
STDC1~\cite{fan2021rethinking}      & \ding{55}          & 170.2                  & 70.2             \\
STDC2~\cite{fan2021rethinking}      & \ding{55}          & 114.0                  & 71.4             \\
\specialrule{0.03em}{0.3em}{0.3em}
DDRNet-23-Slim~\cite{pan2022deep}   & \ding{55}          & 166.7                  & 76.2             \\
DDRNet-23~\cite{pan2022deep}        & \ding{55}          & 109.8                  & 78.2             \\
\specialrule{0.03em}{0.3em}{0.3em}
PIDNet-S~\cite{xu2023pidnet}        & \ding{55}          & 124.9                  & 77.2             \\
PIDNet-M~\cite{xu2023pidnet}        & \ding{55}          & 84.5                   & 78.6             \\
PIDNet-L~\cite{xu2023pidnet}        & \ding{55}          & 64.7                   & 78.7             \\
\specialrule{0.03em}{0.3em}{0.3em}
\rowcolor{green!25}
RDRNet-S-Simple                     & \ding{55}          & 175.1                  & 76.5             \\
\rowcolor{green!25}
RDRNet-S                            & \ding{55}          & 171.5                  & 77.2             \\
\rowcolor{green!25}
RDRNet-M                            & \ding{55}          & 104.0                  & 78.4             \\
\rowcolor{green!25}
RDRNet-L                            & \ding{55}          & 76.1                   & 78.8             \\
\specialrule{0.15em}{0.3em}{0em}
\end{tabular}
}
\label{tab_camvid}
\end{table}

\begin{table}[t]
\renewcommand{\arraystretch}{1.1}
\caption{Accuracy and speed analysis on the Pascal VOC 2012 validation set. All models in the table are results of our reimplementation based on MMSegmentation and were speed-tested on our platform (RTX 3090). During training, all models used Cityscapes weight, and during speed evaluation, convolution and BN were fused. ImageNet indicates whether the model was pre-trained on the ImageNet dataset.}
\centering
\resizebox{\linewidth}{!}
{
\begin{tabular}{lccccc} 
\specialrule{0.15em}{0em}{0.3em}
\textbf{Method}                     & \textbf{ImageNet}  & \textbf{FPS}           & \textbf{PixAcc}  & \textbf{mIoU}          \\ 
\specialrule{0.1em}{0.3em}{0.3em}
ERFNet~\cite{romera2017erfnet}      & \ding{55}          & 58.7                   & 85.6             & 43.4                   \\
\specialrule{0.03em}{0.3em}{0.3em}
ICNet~\cite{zhao2018icnet}          & \ding{55}          & 168.5                  & 87.7             & 50.1                   \\
\specialrule{0.03em}{0.3em}{0.3em}
Fast-SCNN~\cite{poudel2019fast}     & \ding{55}          & 258.7                  & 87.6             & 47.2                   \\
\specialrule{0.03em}{0.3em}{0.3em}
CGNet~\cite{wu2020cgnet}            & \ding{55}          & 99.7                   & 85.6             & 42.2                   \\
\specialrule{0.03em}{0.3em}{0.3em}
BiSeNetV1~\cite{yu2018bisenet}      & \ding{51}          & 124.0                  & 89.5             & 56.0                   \\
BiSeNetV1~\cite{yu2018bisenet}      & \ding{55}          & 124.0                  & 88.9             & 54.2                   \\
BiSeNetV2~\cite{yu2021bisenet}      & \ding{55}          & 139.1                  & 87.7             & 50.0                   \\
\specialrule{0.03em}{0.3em}{0.3em}
STDC1~\cite{fan2021rethinking}      & \ding{55}          & 157.9                  & 88.6             & 52.9                   \\
STDC2~\cite{fan2021rethinking}      & \ding{55}          & 109.9                  & 89.5             & 56.3                   \\
\specialrule{0.03em}{0.3em}{0.3em}
DDRNet-23-Slim~\cite{pan2022deep}   & \ding{55}          & 164.4                  & 88.1             & 53.0                   \\
DDRNet-23~\cite{pan2022deep}        & \ding{55}          & 100.8                  & 89.2             & 56.5                   \\
\specialrule{0.03em}{0.3em}{0.3em}
PIDNet-S~\cite{xu2023pidnet}        & \ding{55}          & 120.4                  & 88.2             & 52.6                   \\
PIDNet-M~\cite{xu2023pidnet}        & \ding{55}          & 77.2                   & 89.3             & 56.2                   \\
PIDNet-L~\cite{xu2023pidnet}        & \ding{55}          & 58.9                   & 89.6             & 56.3                   \\
\specialrule{0.03em}{0.3em}{0.3em}
\rowcolor{green!25}
RDRNet-S-Simple                     & \ding{55}          & 174.2                  & 88.5             & 53.5                   \\
\rowcolor{green!25}
RDRNet-S                            & \ding{55}          & 171.7                  & 88.5             & 53.6                   \\
\rowcolor{green!25}
RDRNet-M                            & \ding{55}          & 96.4                   & 89.7             & 57.2                   \\
\rowcolor{green!25}
RDRNet-L                            & \ding{55}          & 69.7                   & 89.6             & 57.4                   \\
\specialrule{0.15em}{0.3em}{0em}
\end{tabular}
}
\label{tab_pascal}
\end{table}

\subsection{Implementation Details}

\noindent\textbf{Cityscapes.} We utilized Stochastic Gradient Descent (SGD) as the optimizer, with a momentum of 0.9 and a weight decay of 0.0005. The initial learning rate was set to 0.01, and a polynomial learning policy with a power of 0.9 was employed to gradually reduce the learning rate. During training, data augmentation techniques were applied, including random scaling in the range of 0.5 to 2.0, random cropping at a resolution of $1024 \times 1024$, and random horizontal flipping with a probability of 0.5. During inference, the original image with a resolution of $1024 \times 2048$ was used, and no data augmentation was applied. We set the batch size to 12 and trained the model for 120K iterations (about 484 epochs) on two GPUs.

\noindent\textbf{CamVid.} Following the previous works~\cite{pan2022deep, xu2023pidnet}, we employed the Cityscapes pre-trained model and initialized the learning rate to 0.001. During training, we utilized the same data augmentation techniques as Cityscapes, with the exception that the images were randomly cropped to a resolution of $720 \times 960$. During inference, the original image with a resolution of $720 \times 960$ was used. We trained the model for 7800 iterations (200 epochs) on a single GPU, and the remaining hyperparameters were set identically to those used in Cityscapes.

\noindent\textbf{Pascal VOC 2012.} We employed the Cityscapes pre-trained model and set the initial learning rate to 0.001. During training, the image resolution was resized to $512 \times 2048$, and then the same data augmentation techniques as Cityscapes were performed, except that the images were randomly cropped to a resolution of $512 \times 512$. During inference, the image resolution was resized to $512 \times 2048$. It is worth noting that the resolution here was only a target of adjustment. Since the resolution of the images in Pascal VOC is not fixed, the image was adjusted to an approximate resolution based on $512 \times 2048$ in order to maintain the aspect ratio. We trained for 24400 iterations (200 epochs) on two GPUs, and the remaining hyperparameters were set identically to those used in Cityscapes.

\subsection{Computing Platform}

The computing platform hardware we utilize consists of an Intel Xeon Gold 5218R CPU and two NVIDIA RTX 3090 GPUs. The software stack includes Ubuntu 20.04.1, CUDA 11.3, PyTorch 1.12.1, MMSegmentation~\cite{mmseg2020} 1.0.0. During the training phase, we make use of two GPUs, while during the inference phase, only one GPU is employed, with the batch size set to 1.

\subsection{Comparisons with State-of-the-art Models}

We evaluate the performance of our model on three benchmark datasets, namely Cityscapes, CamVid, and Pascal VOC 2012. To ensure a fair comparison, we indicate in the table whether model is pre-trained on the ImageNet~\cite{deng2009imagenet} dataset or not, as some existing works adopt this strategy while others do not. In terms of evaluation metrics, we select mean Intersection over Union (mIoU) as primary metric to assess the performance of all models.

\noindent\textbf{Cityscapes.} The experimental results of the segmentation models on the Cityscapes validation and test sets are presented in Table~\ref{tab_cityscapes}, where the * symbol indicates that we re-trained the model on our platform and evaluated the inference speed of the model on our platform. Notably, the models listed in the table were trained using the combined training and validation sets before being evaluated on the test set. Experimental results show that our RDRNet outperforms other state-of-the-art models in terms of mIoU while maintaining competitive inference speed and model size. As our fastest model, RDRNet-S-Simple achieves 76.8\% mIoU on the validation set, which outperforms other models of the same size, and achieves 134.6 FPS, which also outperforms other models. Although Fast-SCNN is the fastest and achieves 192.0, the performance is only 71.0\%, which is not enough. On the test set, our RDRNet-S-Simple model trails only PIDNet-S. Upon further analysis, we discovered that the $O_{c}$ (this parameter is introduced in \ref{Reparameterizable Dual-Resolution Network}) of the segmentation head in PIDNet-S is 128, whereas our RDRNet-S-Simple has a lower $O_{c}$ of 64, which restricts the model's representation capability. By increasing the $O_{c}$ to 128 (RDRNet-S), our model achieves comparable performance to PIDNet-S on the test set, while still maintaining its speed advantage. Furthermore, our RDRNet-M and RDRNet-L models also demonstrate exceptional performance and fast inference speed compared to other models of similar size. For example, our RDRNet-L outperforms PIDNet-L, with a 0.5\% and 0.2\% mIoU improvement on the validation set and test set, respectively, while achieving an 8.2 FPS speed advantage.

\begin{figure*}[htbp]
\centering
\includegraphics[width=0.98\linewidth]{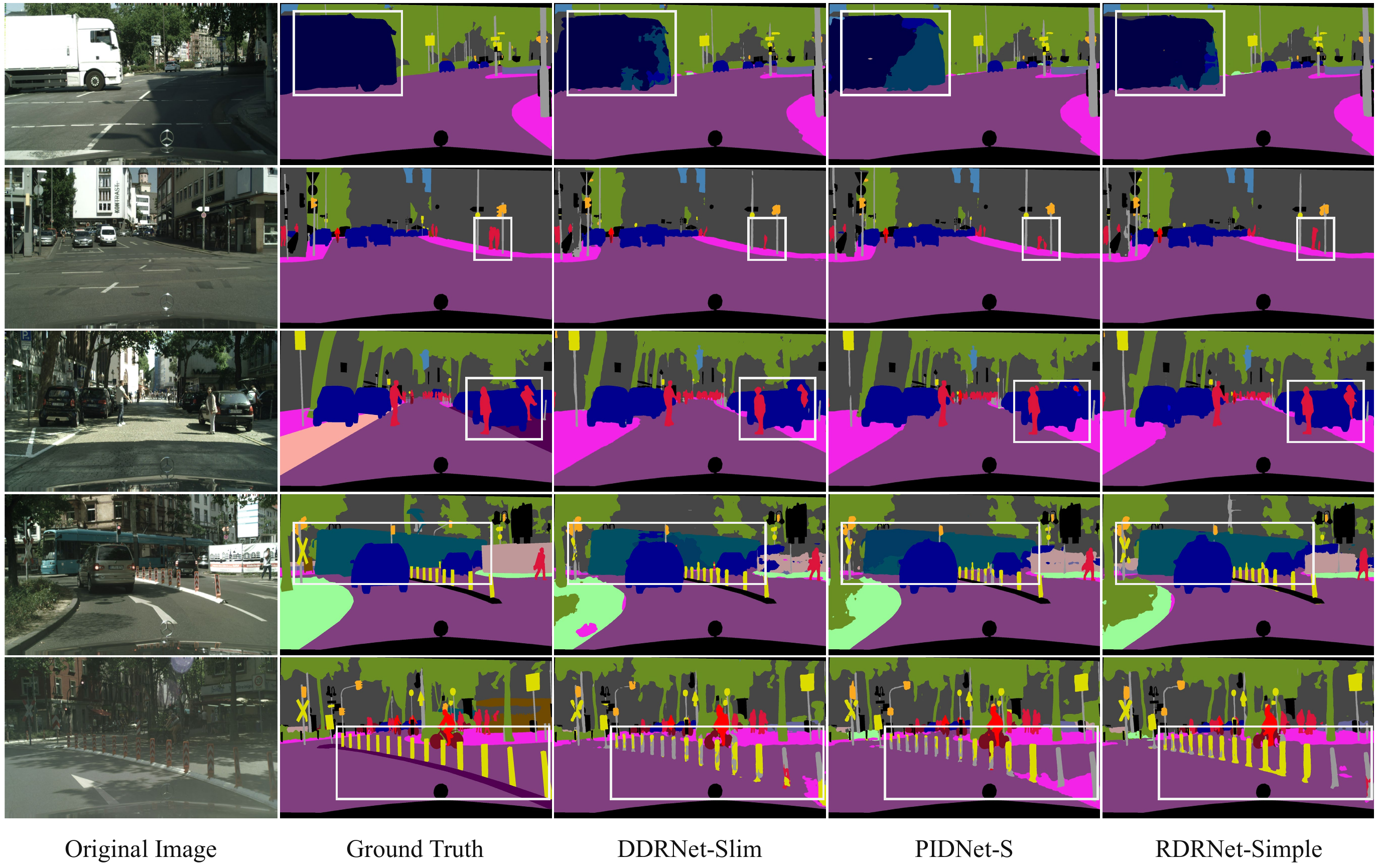}
\caption{Visualization of different segmentation models on the Cityscapes validation set.}
\label{vis_seg}
\end{figure*}

\noindent\textbf{CamVid.} The experimental results of the segmentation models on the CamVid test set are presented in Table~\ref{tab_camvid}, where all models in the table were re-trained on our platform using MMSegmentation, and inference was also performed on our platform. Building upon the previous study~\cite{pan2022deep, xu2023pidnet}, training process leverages a combined dataset consisting of the training set and validation set, while the test set is utilized for model evaluation. During re-training, we reduced the learning rate of all models to $1/10$ of the original learning rate and fine-tuned the Cityscapes pre-trained weight for 7800 iterations, while keeping the other training parameters consistent with those used for Cityscapes. Experimental results demonstrate that our RDRNet maintains a balance between performance and inference speed. When compared to ERFNet, ICNet, CGNet, BiSeNet, and STDC, RDRNet leads in both performance and speed. Although Fast-SCNN is the fastest, its mIoU only reaches 66.2\%. Compared to DDRNet-23-Slim, RDRNet-S-Simple excels in both performance and speed. However, while RDRNet-M achieves a higher mIoU than DDRNet-23, its FPS is slightly lower. When compared to PIDNet-S, RDRNet-S achieves the same mIoU but is 46.6 FPS faster. Additionally, RDRNet-M and RDRNet-L show advantages over PIDNet-M and PIDNet-L, respectively. Specifically, RDRNet-M achieves a 19.5 FPS improvement while maintaining a similar mIoU to PIDNet-M, whereas RDRNet-L outperforms PIDNet-L in mIoU and achieves an 11.4 FPS improvement.

\noindent\textbf{Pascal VOC 2012.} The experimental results of the segmentation models on the Pascal VOC 2012 validation set are presented in Table~\ref{tab_pascal}, where all models in the table were re-trained on our platform using MMSegmentation, and inference was also performed on our platform. During re-training, we reduced the learning rate of all models to $1/10$ of the original learning rate and fine-tuned the Cityscapes pre-trained weight for 24400 iterations, while keeping the other training parameters consistent with those used for Cityscapes. Experimental results indicate that the proposed RDRNet demonstrates an optimal balance between performance and inference speed. Although PIDNet achieves strong results on Cityscapes and CamVid, it struggles to replicate this success on VOC 2012. In contrast, RDRNet consistently delivers excellent performance across all three datasets. As our fastest model, RDRNet-S-Simple outperforms DDRNet-23-Slim and PIDNet-S, respectively, with a 0.6\% and 1.0\% margin in mIoU, and a substantial 9.8 and 53.8 increase in FPS. Additionally, our other RDRNet variants also exhibit strong performance. Beyond the mIoU metric, we supplement our evaluation with the pixel accuracy metric, which is computed by determining the proportion of pixels accurately classified to the total pixel count. Experimental findings indicate that our RDRNet similarly demonstrates superiority in terms of pixel accuracy. Interestingly, the mIoU of BiSeNetV1 on the VOC validation set is particularly outstanding, surpassing BiSeNetV2 by a significant 6.0\%. We are unsure if this is due to the impact of ImageNet pre-training, as BiSeNetV1 was pre-trained on ImageNet but did not perform as well as BiSeNetV2 on the CamVid test set. To verify whether this effect is due to ImageNet pre-training, we conducted additional experiments without using ImageNet pre-training. The experimental results show that without ImageNet pre-training, the mIoU of BiSeNetV1 dropped from 56.0\% to 54.2\%. This indicates that BiSeNetV1 can still perform well on the VOC dataset even without ImageNet pre-training. Although BiSeNetV1 excels on this particular dataset, considering experiments across all three datasets, we conclude that RDRNet outperforms BiSeNetV1 overall.

\begin{table}[t]
\renewcommand{\arraystretch}{1.2}
\setlength{\tabcolsep}{9pt}
\caption{Ablation study of different path combinations of RB on RDRNet-S-Simple.}
\centering
\resizebox{\linewidth}{!}
{
\begin{tabular}{ccccc} 
\specialrule{0.15em}{0em}{0.3em}
\( \mathrm{\mathbf{Conv_{3 \times 3}}} \)   & \( \mathrm{\mathbf{Conv_{1 \times 1}^{(1)}}} \)   & \( \mathrm{\mathbf{Conv_{1 \times 1}^{(2)}}} \)       & \textbf{Residual}  & \textbf{mIoU}   \\ 
\specialrule{0.1em}{0.3em}{0.3em}
\ding{51}                                   &                                                   &                                                       &                    & 75.6            \\
\ding{51}                                   & \ding{51}                                         &                                                       &                    & 76.2            \\
\ding{51}                                   &                                                   &                                                       & \ding{51}          & 76.0            \\
\ding{51}                                   & \ding{51}                                         &                                                       & \ding{51}          & 76.1            \\
\ding{51}                                   & \ding{51}                                         & \ding{51}                                             &                    & 76.7            \\
\ding{51}                                   & \ding{51}                                         & \ding{51}                                             & \ding{51}          & \textbf{76.8}   \\
\specialrule{0.15em}{0.3em}{0em}
\end{tabular}
}
\label{RB ablation experiments 1}
\end{table}

\begin{table}[t]
\renewcommand{\arraystretch}{1.2}
\setlength{\tabcolsep}{23pt}
\caption{Ablation study on the number of $1 \times 1$ convolutions in RB on RDRNet-S-Simple. The training memory refers to the total memory occupied by the model on the two RTX 3090 GPUs during training.}
\centering
\resizebox{\linewidth}{!}
{
\begin{tabular}{ccc} 
\specialrule{0.15em}{0em}{0.3em}
\textbf{Number}                       & \textbf{mIoU}       & \textbf{Training Memory}  \\ 
\specialrule{0.1em}{0.3em}{0.3em}
1                                     & 76.1                & 15.98 GiB                 \\ 
2                                     & 76.8                & 19.27 GiB                 \\ 
3                                     & 76.6                & 22.58 GiB                 \\
\specialrule{0.15em}{0.3em}{0em}
\end{tabular}
}
\label{RB ablation experiments 2}
\end{table}

\begin{figure}[t]
\centering
\includegraphics[width=0.99\linewidth]{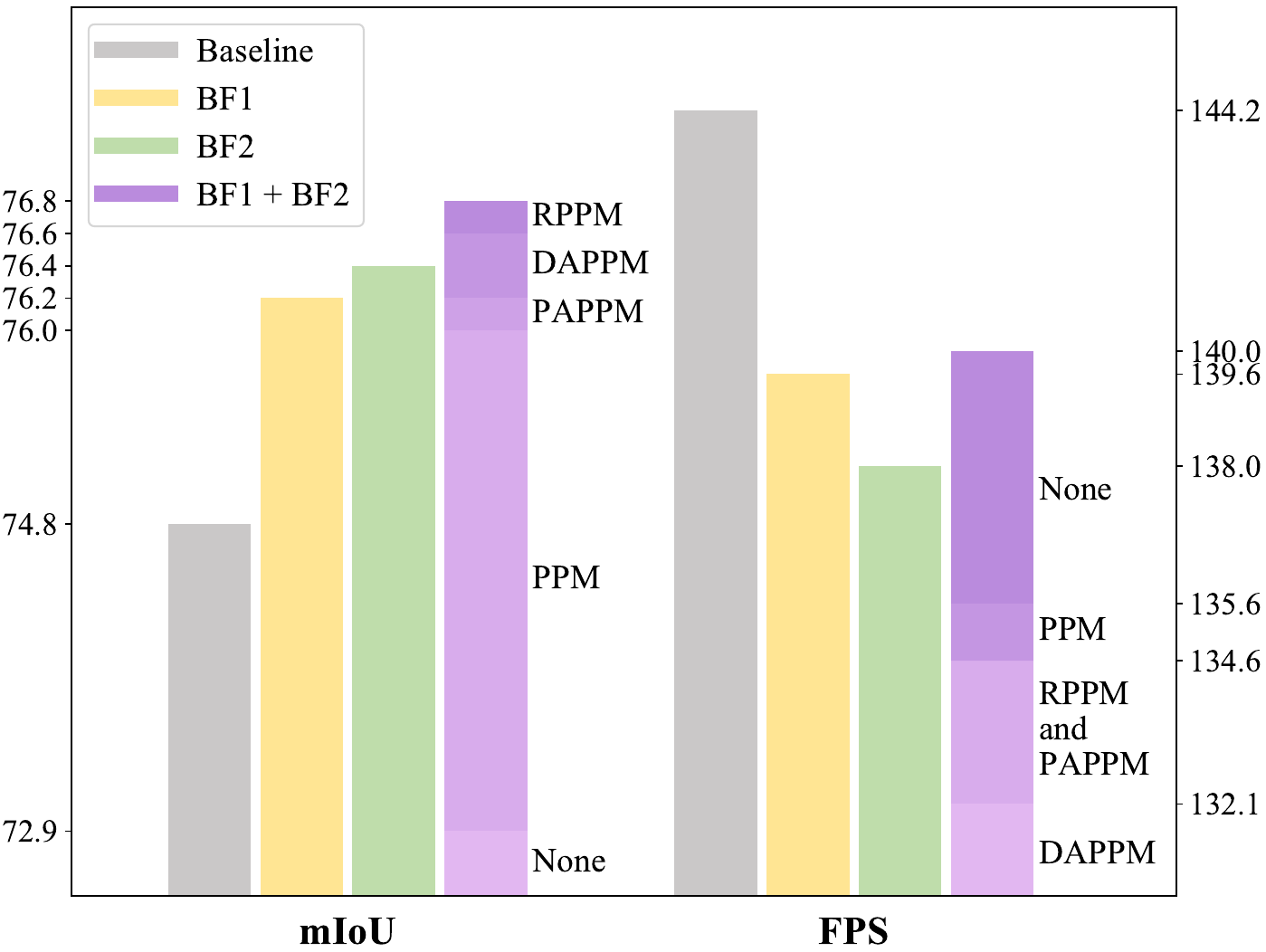}
\caption{Ablation studies of RPPM and bilateral fusion on RDRNet-S-Simple. BF1 denotes bilateral fusion 1 (applied after stage 4), BF2 denotes bilateral fusion 2 (applied after stage 5), and Baseline indicates no bilateral fusion is used. None indicates the absence of the pyramid pooling module. PPM, PAPPM, and DAPPM correspond to \cite{zhao2017pyramid}, \cite{pan2022deep}, and \cite{xu2023pidnet}, respectively. Notably, RPPM is utilized in the first three bars of each group.}
\label{rppm_bf_ablation}
\end{figure}

\subsection{Analysis of Visualization Results}

To clearly demonstrate the performance of our model, we visualized the segmentation results of DDRNet-Slim, PIDNet-S, and our RDRNet-Simple, as shown in Figure~\ref{vis_seg}. The figure shows that RDRNet identifies targets more completely compared to other segmentation models. For example, in the first row, none of the three models fully identify the truck, but RDRNet provides a more complete recognition. The same issue occurs when identifying the bus (fourth row). In the fifth row, DDRNet and PIDNet appear less sensitive to the traffic sign on the ground, whereas RDRNet accurately identifies it. Therefore, RDRNet achieves better segmentation results.

\subsection{Ablation Studies}

\noindent\textbf{Reparameterizable Block.} To verify the effectiveness of RB, we conducted a series of ablation experiments. Specifically, we began the experiment with a single-path block containing a $3 \times 3$ convolution and gradually added a path containing two $1 \times 1$ convolutions and a path with residual connection. Since the $1 \times 1$ convolution path includes two $1 \times 1$ convolutions, we further experimented by reducing these convolutions to explore additional combinations. As shown in Table~\ref{RB ablation experiments 1}, RDRNet's performance was the lowest, with mIoU of just 75.6\%, when using only the single-path block. However, as additional paths were introduced, the performance gradually improved, reaching 76.8\%. Interestingly, the `$\mathrm{Conv_{3 \times 3}}$ + $\mathrm{Conv_{1 \times 1}^{(1)}}$ + Residual' combination performed worse than the `$\mathrm{Conv_{3 \times 3}}$ + $\mathrm{Conv_{1 \times 1}^{(1)}}$' combination, suggesting that residual connection do not always have a positive impact on the model's performance. In addition, since a path containing $1 \times 1$ convolutions can include any number of such convolutions, we performed ablation experiments to determine the optimal number for this path. The experimental results in Table~\ref{RB ablation experiments 2} show that RDRNet performs best when this path includes two $1 \times 1$ convolutions. Each additional $1 \times 1$ convolution increases the memory usage for model training by approximately 3.3 GiB. Notably, we did not test FPS in these experiments because the structure of all combinations remained consistent during inference.

\noindent\textbf{Reparameterizable Pyramid Pooling Module.} To verify the effectiveness of RPPM, we conducted a series of ablation experiments. Specifically, we replaced the RPPM in RDRNet with PPM, DAPPM, and PAPPM, respectively. Additionally, we removed the RPPM entirely, thereby eliminating pyramid pooling module, to observe its impact on model performance. The experimental results are displayed in Figure~\ref{rppm_bf_ablation}, where varying shades of purple represent different modules, and `None' indicates the absence of pyramid pooling module. As shown, the proposed RPPM achieves the highest performance, reaching 76.8\%, which is 0.6\% higher than PAPPM. This performance improvement does not incur additional inference time costs compared to PAPPM, as the structure of RPPM is reparameterized to be consistent with that of PAPPM during inference. When pyramid pooling module is not used, the model achieves the fastest inference speed, reaching 140.0 FPS, but the mIoU is only 72.9\%. We believe it is reasonable to sacrifice 5.4 FPS for a 3.9\% improvement in mIoU.

\noindent\textbf{Bilateral Fusion.} To verify the effectiveness of two bilateral fusion, we conducted a series of ablation experiments. Specifically, we delve deeper into the scenarios where RDRNet does not utilize bilateral fusion, where RDRNet only employs bilateral fusion 1 (applied after stage 4), and where RDRNet only utilizes bilateral fusion 2 (applied after stage 5). Figure~\ref{rppm_bf_ablation} shows that without bilateral fusion, RDRNet achieves the lowest mIoU at 74.8\%, yet attains the fastest FPS, reaching 144.2. Compared with utilizing bilateral fusion 1 alone, employing bilateral fusion 2 alone can achieve higher performance, albeit with a slower inference speed. When both bilateral fusion 1 and bilateral fusion 2 are utilized simultaneously, the mIoU reaches its peak at 76.8\%, albeit with a corresponding drop in FPS to 134.6. All in all, employing two bilateral fusion proves effective for enhancing performance, despite the reduction in the model's inference speed, which remains acceptable.

\section{Conclusion}

In this study, we propose a Reparameterizable Dual-Resolution Network (RDRNet) for real-time semantic segmentation. By utilizing multi-path blocks during training and reparameterizing them to single-path blocks during inference, we optimize both accuracy and speed. Additionally, we introduce a Reparameterizable Pyramid Pooling Module (RPPM) to enhance feature representation without increasing inference time. Extensive experimental results demonstrate that RDRNet outperforms existing state-of-the-art models, offering both high performance and fast inference capabilities. In the future, we plan to explore more potent reparameterizable training structures.

\bibliographystyle{IEEEtran}
\bibliography{sample}

\end{document}